\documentclass[times,twocolumn,final]{elsarticle}

\usepackage{ipcai_arxiv}
\usepackage{framed,multirow}

\usepackage{amssymb}
\usepackage{latexsym}

\usepackage{url}
\usepackage{cite}
\usepackage{amsmath,amssymb,amsfonts,bm}
\usepackage{booktabs}
\usepackage{arydshln}
\usepackage{tabulary}
\usepackage{graphicx,epstopdf}
\usepackage{textcomp}
\usepackage{subcaption}
\usepackage{booktabs}
\usepackage[svgnames,table]{xcolor}
\usepackage{pifont}
\usepackage{arydshln}
\usepackage[colorlinks=true,citecolor=cyan,urlcolor=cyan]{hyperref}

\usepackage{placeins}
\usepackage{fancyhdr}

\fancypagestyle{firstpagestyle}{
    \fancyhf{} 
    \fancyhead{} 
    \fancyfoot{} 
    \fancyhead[CO]{\em \fontsize{9pt}{8pt}\selectfont This article has been accepted to the International Conference on Information Processing in Computer-Assisted Interventions, 2024.}
}

\fancypagestyle{default}{
    \fancyhf{}
    \fancyhead[R]{\thepage} 
    \fancyhead[LO]{\em \fontsize{9pt}{8pt}\selectfont This article has been accepted to the International Conference on Information Processing in Computer-Assisted Interventions, 2024.}
}

\definecolor{newcolor}{rgb}{.8,.349,.1}

\def\maketitlesupplementary
   {
   \newpage
       \twocolumn[
        \centering
        \Large
        \textbf{Advancing Surgical VQA with Scene Graph Knowledge}\\
        \vspace{0.5em}Supplementary Material \\
        \vspace{1.0em}
       ]
   }

\begin{document}


\begin{frontmatter}

\title{Advancing Surgical VQA with Scene Graph Knowledge}

\author[1,3]{Kun \snm{Yuan}}
\author[1,2]{Manasi \snm{Kattel}}
\author[2]{Jo\"el L. \snm{Lavanchy}}
\author[3]{Nassir \snm{Navab}}
\author[1,2]{Vinkle \snm{Srivastav}}
\author[1,2]{Nicolas \snm{Padoy}}
\cortext[cor1]{Corresponding author: 
  Tel.: +33-390413530}
\ead{npadoy@unistra.fr}

\address[1]{ICube, University of Strasbourg, CNRS, Strasbourg, France}
\address[2]{IHU Strasbourg, Strasbourg, France}
\address[3]{CAMP, Technische Universit\"at M\"unchen, Munich, Germany}

\received{XXX}
\finalform{XXX}
\accepted{XXX}
\availableonline{XXX}
\communicated{XXX}

\begin{abstract}

\textbf{Purpose:}
The modern operating room is becoming increasingly complex, requiring innovative intra-operative support systems. While the focus of surgical data science has largely been on video analysis, integrating surgical computer vision with natural language capabilities is emerging as a necessity. Our work aims to advance Visual Question Answering (VQA) in the surgical context with scene graph knowledge, addressing two main challenges in the current surgical VQA systems: removing question-condition bias in the surgical VQA dataset and incorporating scene-aware reasoning in the surgical VQA model design.

\textbf{Methods:}
First, we propose a \textbf{S}urgical \textbf{S}cene \textbf{G}raph-based dataset, SSG-VQA, generated by employing segmentation and detection models on publicly available datasets. We build surgical scene graphs using spatial and action information of instruments and anatomies. These graphs are fed into a question engine, generating diverse QA pairs. Our SSG-VQA dataset provides a more complex, diverse, geometrically grounded, unbiased, and surgical action-oriented dataset compared to existing surgical VQA datasets. We then propose SSG-VQA-Net, a novel surgical VQA model incorporating a lightweight Scene-embedded Interaction Module (SIM), which integrates geometric scene knowledge in the VQA model design by employing cross-attention between the textual and the scene features. 

\textbf{Results:}
Our comprehensive analysis of the SSG-VQA dataset shows that SSG-VQA-Net outperforms existing methods across different question types and complexities. We highlight that the primary limitation in the current surgical VQA systems is the lack of scene knowledge to answer complex queries. 

\textbf{Conclusion:}
We present a novel surgical VQA dataset and model and show that results can be significantly improved by incorporating geometric scene features in the VQA model design. The source code and the dataset will be made publicly available at: \url{https://github.com/CAMMA-public/SSG-VQA}
\end{abstract}

\end{frontmatter}
\thispagestyle{firstpagestyle}

\section{Introduction}
\label{sec1}

Surgical data science is a rapidly growing field that aims to streamline clinical workflows and enable the development of real-time intra-operative decision-support systems~\citep{maier2022surgical,padoy2019machine}. Recent advancements in surgical video analysis, such as surgical workflow phase recognition, fine-grained surgical action detection, and surgical semantic scene segmentation, show evidence of the progress ~\citep{nwoye2021rendezvous,carstens2023dresden,twinanda2016endonet}. However, the scope of these methods is confined as it mainly focus on visual-only data to perform classification or recognition tasks, thereby offering limited user interaction. The next generation of surgical data science applications also demands approaches operating at the crucial intersection of vision and language to offer intuitive user interaction during intra-operative surgical procedures. Surgical Visual Question Answering (VQA) is emerging as a notable solution in that direction, which aims to provide precise answers to user queries in a natural language by analyzing a given surgical image ~\citep{antol2015vqa,hudson2019gqa,seenivasan2022surgical,seenivasan2023surgicalgpt}.

\begin{figure*}[]
\includegraphics[width=\textwidth]{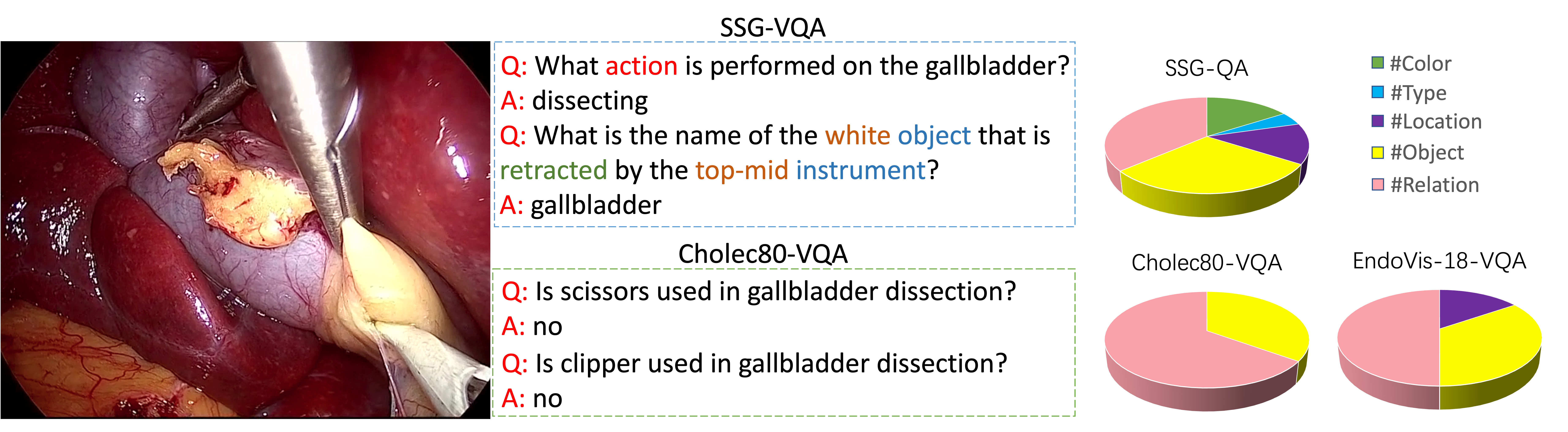}
\caption{SSG-VQA dataset contains up to $50$ complex visual reasoning questions, compared to $2$ classification-based questions in the Cholec-VQA ~\citep{seenivasan2022surgical}.} \label{fig1}
\end{figure*}

Developing an effective surgical VQA system is inherently challenging for a typical surgical scene, which contains multiple anatomical structures and instruments connected through diverse spatial and action relationships. While a few works have explored VQA tasks in the surgical context ~\citep{seenivasan2022surgical,seenivasan2023surgicalgpt}, they are typically limited to datasets and models that ignore detailed scene knowledge. From the dataset perspective, one key challenge is the lack of a dataset with potentially a vast set of question-answer pairs covering various aspects of the surgical scene. The current surgical VQA datasets are small and only consider simple scene information, e.g., object/action occurrence, as shown in Fig.~\ref{fig1}. Moreover, these datasets contain question-answer pairs with significant question-condition bias, where answers can be derived from just the questions without performing any visual processing. This hinders the utility of these datasets to serve as appropriate surgical VQA benchmarks.

From the model perspective, the current surgical VQA architectures operate on the global visual representation of the surgical image, ignoring the detailed understanding of surgical scene knowledge. This can be detrimental, especially when object-level visual reasoning is essential to answer fine-grained questions. Our key contributions are therefore twofold: the introduction of a new surgical scene-aware VQA dataset called SSG-VQA and a novel surgical VQA model called SSG-VQA-Net. 

The SSG-VQA dataset uses a semantic scene graph ~\citep{krishnavisualgenome} as a suitable representation to generate diverse question-answer pairs. A semantic scene-graph representation provides scene knowledge by detecting objects and their attributes and connecting relationships and interactions between objects in the scene. To develop the surgical semantic scene graph, we use publicly available datasets for semantic segmentation and tool detection ~\citep{jin2018tool, hong_kao_kuo_wang_chang_shih_2020}, and apply these models to estimate object spatial relationships. We then estimate the surgical action relationships, i.e., $<instrument, verb, target>$, among the objects using the public CholecT45 dataset ~\citep{nwoye2021rendezvous}.

We then develop a surgery-specific question engine that ingests the surgical scene graph and manually designed question templates to produce a variety of question-answer pairs. Including detailed surgical scene understanding along with question templates helps us to generate question-answer pairs covering various aspects of the surgical scene, for example, fine-grained action recognition - ``What is the action being performed on peritoneum?'', semantic scene reasoning - ``What anatomy is at the top-mid of the frame?'',  and surgical object attribute reasoning - ``What is the name of the anatomy that is being retracted?''. Furthermore, we perform the balancing and sampling strategies based on the surgical-specific knowledge and class distribution to remove the questions that contain question-condition bias, e.g., ``How many livers are in the frame?'' which counts the number of certain anatomical structures. The overall pipeline is illustrated in Fig.~\ref{fig2}. 

Given a large-scale SSG-VQA dataset containing fine-grained surgical question-answer pairs, we propose a multi-modality surgical VQA model called SSG-VQA-Net. Existing surgical VQA models use a highly parameterized multi-modality transformer encoder to fuse the textual embeddings from a question and the patches from a global visual representation of a surgical image ~\citep{seenivasan2022surgical,seenivasan2023surgicalgpt}. However, these patches do not contain object-wise information about the surgical scene, hence missing the geometric scene understanding. Our key idea is to exploit object-wise local features and fuse geometric scene information in the VQA model design. To enable this, we train a fast and lightweight object detector, YOLOv7 ~\citep{wang2023YOLOv7}, on the bounding-box labels of the SSG-VQA dataset. The trained object detector allows us to extract object-wise local representations of the surgical scene objects using RoIAlign pooling ~\citep{he2017mask}. Furthermore, we integrate the geometric spatial coordinates and class labels of detected bounding boxes into the VQA model by introducing a lightweight multi-modal transformer encoder named the Scene-embedded Interaction Module (SIM). The SIM module uses a scene graph of detected bounding boxes where each node contains the class label and bounding-box coordinate information. The scene graph is refined by cross-attention between the scene graph and the textual inputs, highlighting specific graph nodes correlated to the complex question query. These refined text-aware scene embeddings are then combined with the object-wise local representations of the surgical scene and the textual embeddings through a transformer encoder layer to generate an accurate response. Experimental results show that our method outperforms prior works while achieving a low parameter count. We summarize our main contributions as follows:

\begin{itemize}
    \item We present a new surgical scene graph-based VQA dataset, SSG-VQA, providing complex, diverse, geometrically grounded, and surgical action-oriented question answers.
    \item We present a surgical VQA model, SSG-VQA-Net, utilizing a novel scene-aware feature extraction strategy to obtain state-of-the-art performance.
\end{itemize}

\thispagestyle{default}

\section{Methodology}

\subsection{SSG-VQA dataset}
\label{sec:dataset}

In this section, we explain the SSG-VQA dataset generation process consisting of creating surgical scene graphs, designing a question engine with diverse templates, and employing a sampling strategy to mitigate data imbalance and question-condition bias.

\subsubsection{Scene graph generation}

We build our SSG-VQA dataset using the publicly available CholecT45 ~\citep{nwoye2021rendezvous}, m2cai16-tool-locations ~\citep{jin2018tool} and CholecSeg8k ~\citep{hong_kao_kuo_wang_chang_shih_2020} datasets. Specifically, we train a tool detection model ~\citep{ultralytics} on m2cai16-tool-locations ~\citep{jin2018tool} and a semantic segmentation model ~\citep{deeplabv3plus2018} on CholecSeg8k ~\citep{hong_kao_kuo_wang_chang_shih_2020} to extract bounding boxes of surgical objects, including surgical instruments and anatomies. Then, we build the surgical semantic scene graph using the detections. A surgical scene graph can be formulated as a set of nodes and edges, where the nodes represent surgical objects that contain a set of attributes, i.e., color, location, and type, and edges represent the spatial and action relations among the objects. The spatial relations are calculated by comparing the centroid of objects, and the action relations are provided by the triplet annotations from CholecT45 ~\citep{nwoye2021rendezvous}. Then, we leverage the generated scene graphs as input to a question engine, as described below, to generate diverse question-answer pairs. Note that to create a clean test set of question-answer pairs, we manually correct the bounding boxes and class labels of scene graphs in the test videos.

\begin{figure*}[t!]
\centering
\includegraphics[width=1\textwidth]{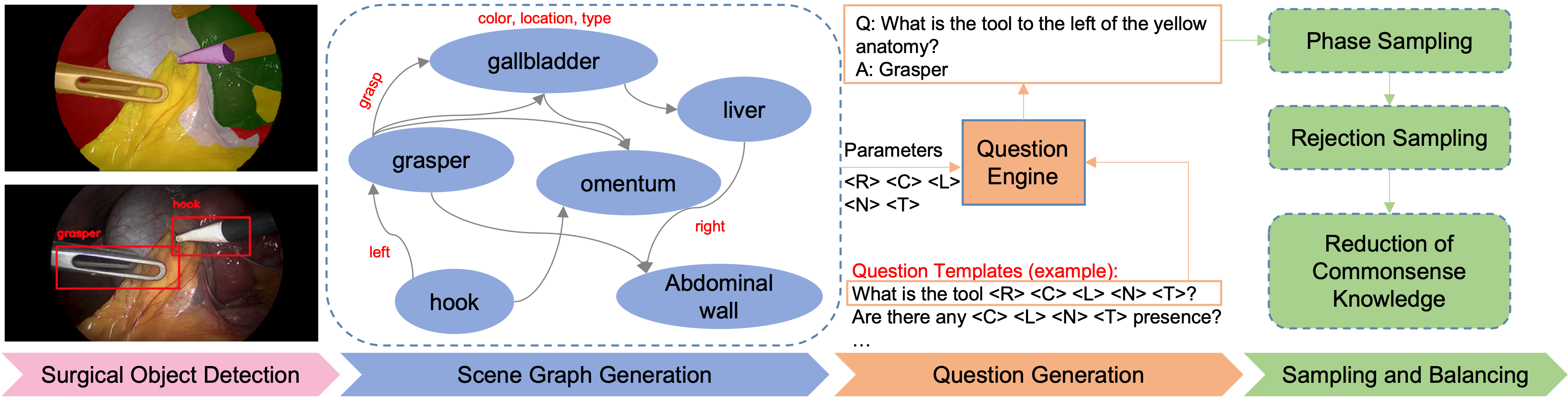}
\caption{Pipeline of SSG-VQA construction. The dataset is constructed from the well-designed question engine, which takes the scene graph as input and changes the parameters of question templates to generate diverse question-answer pairs.} \label{fig2}
\end{figure*}

\subsubsection{Question engine}

The question engine, which is responsible for generating diverse questions with various categories, requires two inputs, i.e., scene graph and question templates. We use the CLEVR engine ~\citep{johnson2017clevr,hudson2019gqa} and extend it to the surgical context. Specifically, the question engine can change question templates' parameters conditioned on the surgical scene graph to express diverse questions. For example, the question ``what is the tool to the left of yellow anatomy?'' can be formed by the template ``what is the tool $<$R$>$ $<$C$>$ $<$L$>$ $<$T$>$?'', by replacing the parameters $<$R$>$, $<$C$>$, $<$L$>$ and $<$T$>$ into ``to the left of'', ``yellow'', ``null'' and ``anatomy''. The questions are parameterized by five parameters, namely $<$C$>$ (color), $<$L$>$ (location), $<$T$>$ (type), $<$N$>$ (name), and $<$R$>$ (relationship). In total, there are $40$ question templates containing different types of questions, such as \textit{querying object} (e.g., ``what is the name of instruments to the left of the gallbladder?''), \textit{querying attribute} ( e.g., ``there is an object that is both to the left of the yellow thing and below the brown anatomy; what is its location?''), \textit{querying relation} (e.g., ``what is the action being performed?''), \textit{confirming existence} (e.g., ``is there a bipolar in the top-mid location?''), and \textit{counting} (e.g., ``how many instruments are in the bottom-right?''). 

Generated questions also fall into three categories depending on their complexity: \textit{zero-hop}, \textit{one-hop}, and \textit{single-and}. Each requires different visual reasoning steps to resolve. Specifically, solving these three types of questions involves the understanding of relations between zero, one, or two surgical objects, respectively. Examples from each category are provided in the supplementary material. The question engine allows us to freely control the complexity, length, and number of questions per image. 

\subsubsection{Sampling and balancing}
Here, we introduce the applied strategies to reduce the effect of class imbalance and question-condition bias during SSG-VQA dataset construction. Surgical VQA is a multi-choice task, which mainly includes questions about the surgical objects. Therefore, an imbalance in the occurrence of surgical objects could lead to an imbalance in their class distribution. To address that, we sample the frame amounts based on the surgical phase and tool presence labels from the Cholec80 dataset, instead of sampling evenly like Cholec80-VQA ~\citep{seenivasan2022surgical}. Also, to address the question-condition bias that the information is leaked out from poorly formulated questions, we remove the question templates that may contain the question-condition bias. We also eliminate poorly formulated questions, such as ``What is the location of the $<$N$>$?'' with $<$N$>$=gallbladder when there is no gallbladder in the scene. These processing strategies prevent question-condition bias and avoid generating degenerate question-answer pairs. The overall pipeline is shown in Fig.~\ref{fig2}.

\subsection{SSG-VQA-Net}
\label{sec:model}

\subsubsection{Pipeline}

\begin{figure*}[t!]
\centering
\includegraphics[width=0.9\textwidth]{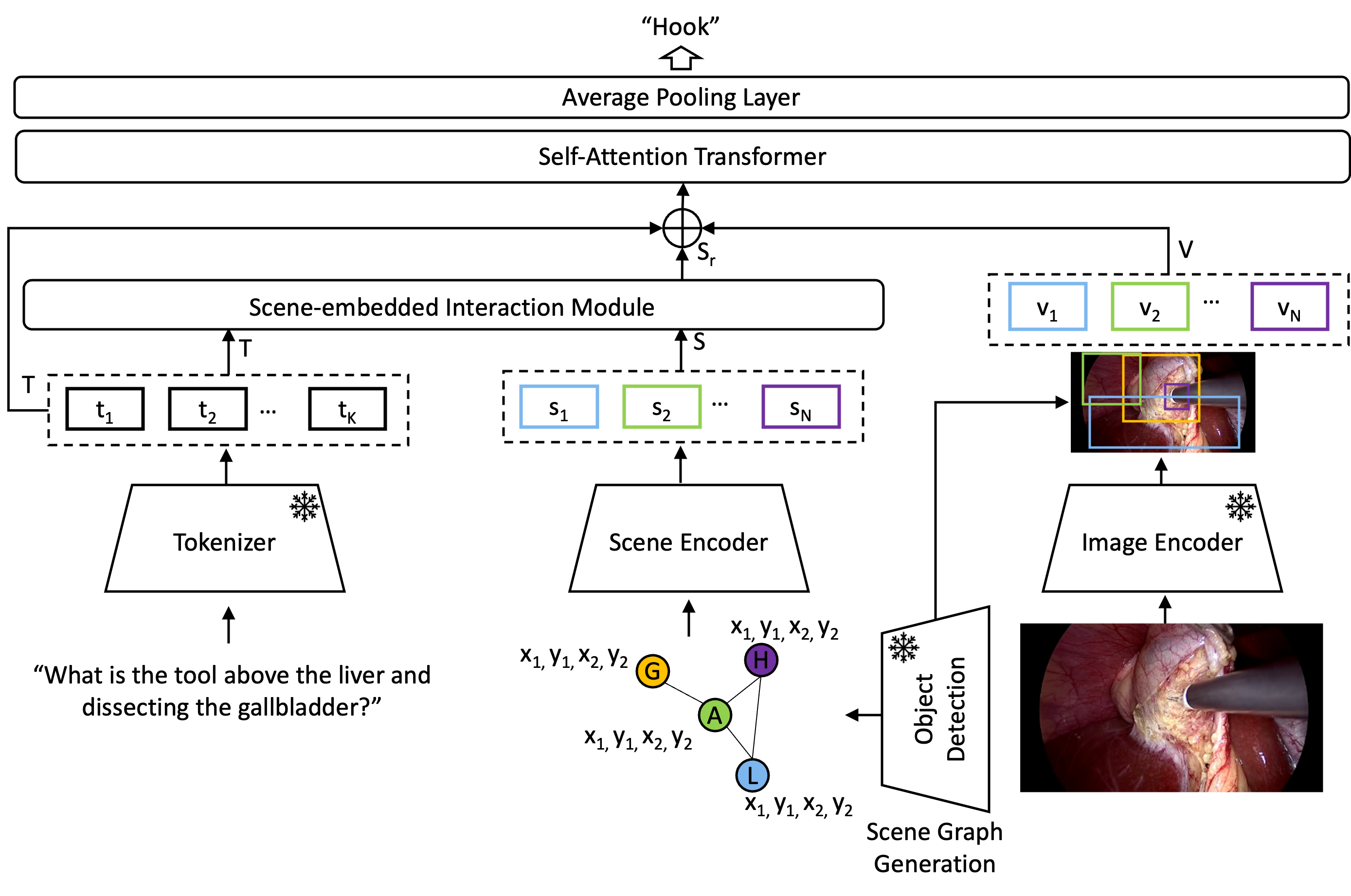}
\caption{Pipeline of the SSG-VQA-Net. It requires three types of inputs, textual, visual, and scene knowledge. The textual and scene embeddings are fed into the SIM and generate refined scene embeddings. The visual embeddings are generated from the RoIAlign. Finally, we concat them to feed into the self-attention transformer to get the final answer. Here, G, H, A, and L represent class labels; $x_1$, $y_1$, $x_2$, and $y_2$ represent bounding box coordinates (G: gallbladder, H: hook; A: abdominal wall cavity; L: Liver).} \label{fig:our_model}
\end{figure*}

Here, we explain the pipeline of SSG-VQA-Net. Given the textual form of a question, we first extract textual embeddings of questions using a pre-trained tokenizer ~\citep{seenivasan2022surgical}, denoted as $T=\{t_1,...t_{K}\}$. From the surgical scene image, we extract a feature map using the ResNet18 ~\citep{resnet18} visual backbone. Then we use a trained object detector, YOLOv7 ~\citep{wang2023YOLOv7}, to detect the surgical objects and extract $N$ object-wise visual embeddings using RoIAlign pooling, denoted as $V=\{v_1,...v_{N}\}$. The object detector is trained on the bounding boxes of surgical objects from the SSG-VQA training dataset. 

We build the scene embeddings using the detected surgical objects' information. Specifically, we initialize the graph nodes as a concatenation of objects' class labels and spatial coordinates, as shown in Fig.~\ref{fig:our_model}. These low-dimensional embeddings are projected using a linear layer, called Scene Encoder, to match the dimensionality of textual embeddings. These scene embeddings  $S=\{s_1,...s_{N}\}$ are then passed through our proposed Scene-embedded interaction (SIM) module, explained below, to obtain text-aware scene embeddings ($S_r$). These text-aware scene embeddings ($S_r$) are then concatenated with the visual embeddings ($V$) and the textual embeddings ($T$) and passed through a self-attention-based transformer module. Finally, features are average-pooled and mapped to a predefined answer set to generate the output answer.

\thispagestyle{default}

\subsubsection{Scene-embedded interaction Module}

\begin{figure*}[t!]
\centering
\includegraphics[width=0.8\textwidth]{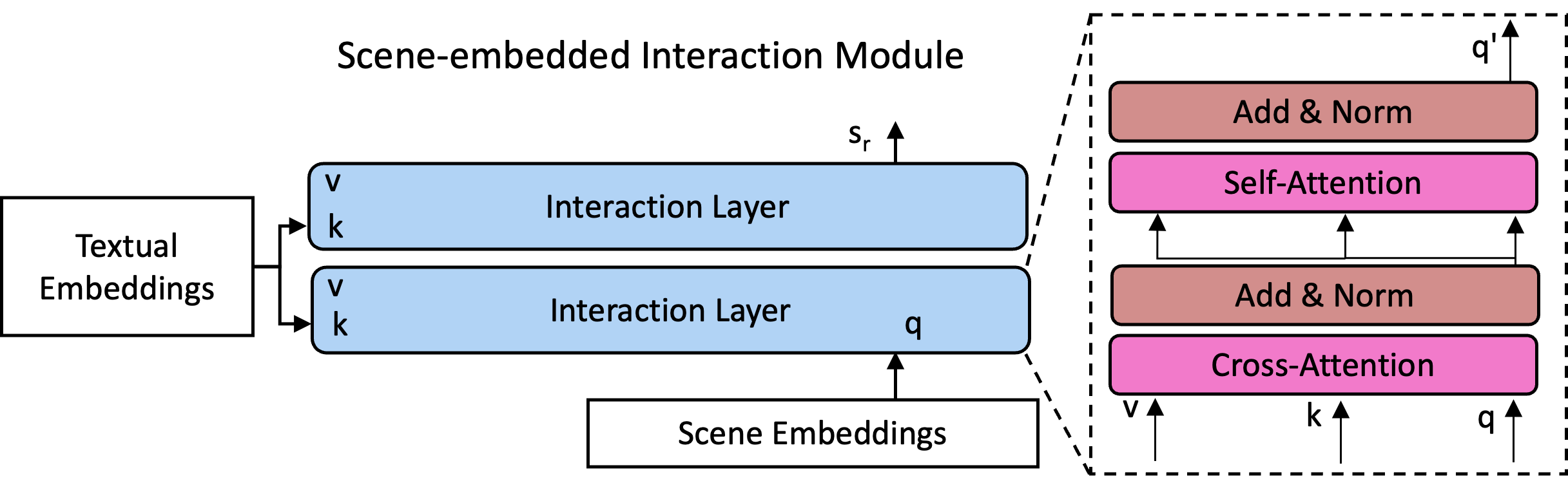}
\caption{Scene-embedded interaction module. It is a stack of layers of cross-attention and self-attention. The cross-attention modulates the scene embeddings based on the text queries, while the self-attention refines the scene embeddings .} \label{fig4}
\end{figure*}

In SSG-VQA-Net, initial scene embeddings $S$ capture global surgical scene semantics. To handle complex questions that require localized focus, we introduce a lightweight Scene-embedded Interaction Module (SIM). 
The main objective of SIM is to correlate the textual embeddings with the scene embeddings. SIM consists of two interaction layers. Each layer comprises self-attention, cross-attention, and feed-forward sub-layers, as shown in Fig.~\ref{fig4}. The attention mechanism is defined as:
\begin{equation}
    \text{Attention}(Q, K, V) = \text{softmax}\left(\frac{QK^T}{\sqrt{d_k}}\right) V, 
\end{equation}
where the Q is the short for query, K is key, and V is value.

In SIM, we first apply cross-attention, $S_r = \text{Cross-Attention}(S, T, T)$, to the textual and scene embeddings by processing textual embeddings $T$ as key and value, and scene embeddings $S$ as the query. This results in refined scene embeddings, which incorporate textual cues. Then, the refined scene embeddings are passed to the self-attention layer, $S_r = \text{Self-Attention}(S_r, S_r, S_r)$, to interact with themselves. By interacting with the textual embeddings and the scene embeddings, we obtain the textual-aware scene embeddings $S_r$. Through our ablation experiments, we show that the $S_r$ significantly contributes to providing correct answers to fine-grained questions.

\thispagestyle{default}

\section{Results and discussions}

\begin{table*}[t!]
\centering
\caption{We show that the SSG-VQA dataset is more challenging and balanced than the EndoVis-18-VQA~\citep{seenivasan2022surgical} and Cholec80-VQA~\citep{seenivasan2022surgical}. SSG-VQA dataset includes more attributes and complexities in the questions.}
\label{table:dataset_stats}
\begin{tabular}{c@{\hspace{0.5cm}}c@{\hspace{0.5cm}}c@{\hspace{0.5cm}}c} 
\toprule
Dataset & EndoVis-18-VQA ~\citep{seenivasan2022surgical} & Cholec80-VQA ~\citep{seenivasan2022surgical} & SSG-VQA (Ours)  \\
\midrule
$\#$Surgical Scenes    & 2k & 21k & 25k\\ 
$\#$Questions & 11k & 43k & 960k \\
$\#$Unique Questions & 17 & 51 & 501k \\
\midrule
Average Length in words & 5.8  & 2.0 & 12.8 \\
Average $\#$Questions per scene & 5.0 & 6.5 & 38.9
\\\bottomrule\hline
\end{tabular}
\end{table*}

\subsection{Dataset comparison}
\label{sec:dataset_comparison}

SSG-VQA dataset contains the same train and test set videos as CholecT45 ~\citep{nwoye2021rendezvous} dataset, which contains $40$ laparoscopic cholecystectomy videos for training and $5$ videos for testing. Our SSG-VQA dataset contains $960k$ questions from $25k$ surgical scenes. Table ~\ref{table:dataset_stats} presents the comparison between SSG-VQA and the typical datasets from prior work, i.e., EndoVis-18-VQA and Cholec80-VQA from ~\citep{seenivasan2022surgical}, showing a $8\times$ and $22\times$ increase in number of questions, respectively. Also, our SSG-VQA dataset contains more diverse questions per scene ($38.9$ v.s. $6.5$) and much longer questions ($12.8$ words v.s. $5.8$ words). Furthermore, SSG-VQA contains a wider range of categories for object attributes, names, and inter-object relationships, as shown in the appendix. Also, compared to the Cholec80-VQA dataset which provides $51$ questions for all surgical scenes, our SSG-VQA has more diverse questions ($501k$) that are unique to surgical scenes, which prevents the model from overfitting to specific question patterns.

\subsection{Question-condition bias}
\label{sec:question_condition}

\begin{table*}[t!]
\centering
\caption{Identification of question-condition bias in existing datasets. We use the Accuracy, Recall, and Fscore metrics from SurgicalVQA ~\citep{seenivasan2022surgical}. Endovis-VQA contains significant question-condition bias because the language model with pure language inputs can outperform the model with vision and language inputs.}
\label{table:question-condition bias}
\begin{tabular}{c@{\hspace{0.3cm}}c@{\hspace{0.3cm}}c@{\hspace{0.3cm}}c@{\hspace{0.3cm}}c@{\hspace{0.3cm}}c@{\hspace{0.3cm}}c@{\hspace{0.3cm}}c@{\hspace{0.3cm}}c@{\hspace{0.3cm}}c} 
\toprule
Methods & \multicolumn{3}{c}{Endovis-VQA ~\citep{seenivasan2022surgical}} & \multicolumn{3}{c}{Cholec80-VQA ~\citep{seenivasan2022surgical}} & \multicolumn{3}{c}{SSGQA} \\
\hline
& Acc&Recall&Fscore & Acc&Recall&Fscore &  Acc&Recall&Fscore \\
 \midrule
L+Bert ~\citep{devlin2018bert} & 57.5&45.9&36.3 & 83.3&29.3&24.4 & 51.7 & 36.2 &50.0 \\
L+SciBert ~\citep{beltagy2019scibert} & 55.8&45.1&36.9 & 83.3&29.3&24.4 & 52.4 & 36.9 & 49.4\\
L+ClinicalBert ~\citep{huang2019clinicalbert} & 60.4&\textbf{50.7}&\textbf{40.8} & 83.1&29.4&24.4 & 52.3 &35.4 & 49.9\\\hline
VisualBert ~\citep{visualbert} & 61.9&41.2&33.4 & 89.7&\textbf{62.9} &63.3 & 55.0 &42.5 &54.8 \\
VisualBert Resmlp ~\citep{visualbert} & \textbf{63.2}&39.6&33.6 & \textbf{89.8}&62.7&\textbf{63.4} & \textbf{58.7} & \textbf{44.8} & \textbf{57.7} \\
\bottomrule
\end{tabular}

\end{table*}

VQA systems can exploit the question-condition bias from the dataset as a shortcut to answer questions without understanding the visual scenes. To quantify this bias, we train language-only models like ClinicalBert ~\citep{huang2019clinicalbert} to answer the questions without any visual information on existing VQA datasets, such as EndoVis-18-VQA and Cholec80-VQA, and on our SSG-VQA dataset. As shown in Table~\ref{table:question-condition bias}, the language-only model ClinicalBert outperforms the vision-language multi-modal models in EndoVis-18-VQA, suggesting the questions from EndoVis-18-VQA contain simple shortcuts to the correct answer. Cholec80-VQA and SSG-VQA have a lower bias as their questions are more vision-relevant. SSG-VQA further reduces bias by using scene-graph-based diverse questions. In the following, we perform the experiments on the Cholec80-VQA and SSG-VQA dataset due to their low question-condition bias.

\begin{table*}[t!]
\centering
\caption{Comparison results for baselines and our models. The SSG-VQA-Net with scene knowledge achieves the best results. The SSG-VQA-Net (oracle) model refers to the model that uses detection labels from the SSG-VQA dataset to construct the scene embeddings instead of using the trained YOLOv7 object detector.}\label{table:overall}
\begin{tabular}
{c@{\hspace{1.2cm}}c@{\hspace{1.2cm}}c@{\hspace{1.2cm}}c@{\hspace{1.2cm}}c@{\hspace{1.2cm}}c} 
\toprule
  Models   & Accuracy & mAP & Recall & Fscore  \\
\midrule
L+ClinicalBert & 52.3 & 40.4 & 35.4 & 49.9  \\ 
VisualBert ~\citep{visualbert} & 55.0 & 47.9 & 42.5 & 54.8 \\
VisualBert Resmlp ~\citep{seenivasan2022surgical}   & 58.7 & 51.8 & 44.8 & 57.7 \\
SurgicalGPT~\citep{seenivasan2023surgicalgpt} & 58.5 & 49.4 & 44.8 & 57.8 \\
SSG-VQA-Net & 60.7 & 54.9 & 49.1 & 60.3 \\ 
SSG-VQA-Net (Oracle) & \textbf{62.8} & \textbf{56.3} & \textbf{50.6} & \textbf{62.3} \\ 

\bottomrule

\end{tabular}
\end{table*}

\begin{table*}[h!]
\centering
\caption{Breakdown results of the prior models and our models. We show that our model outperforms the baselines by a large margin, especially on the complex questions that require visual reasoning. Also, the results on a different set of questions shows that our dataset is not dominated by one type of question. We report the mAP here.}\label{table:ssgqa_breakdown}
\begin{tabular}{c@{\hspace{0.28cm}}c@{\hspace{0.28cm}}c@{\hspace{0.28cm}}c@{\hspace{0.28cm}}c@{\hspace{0.28cm}}c} 
\toprule
& VisualBERT ~\citep{visualbert} & VisualBERTMLP ~\citep{seenivasan2022surgical} & SSG-VQA-Net & SSG-VQA-Net (oracle)  \\ \midrule
Query Object & 39.4  & 38.4   & 48.0 & \textbf{55.4}\\
Query Attribute   & 51.7 & 54.5 & 54.8 & \textbf{60.2} \\
Existence    & 68.0 & \textbf{76.4}   & 73.9 & 72.7 \\
Counting    & 24.5 & 29.6 & \textbf{36.9} & 24.2 \\  
\hline
Zero-hop & 50.4 & 53.2 & \textbf{56.6} & 55.0 \\
One-hop & 46.4 & 46.2 & 50.3  & \textbf{51.9} \\
Single-and & 23.4 & 30.9 & 39.0 & \textbf{41.4} \\ 
\bottomrule
\end{tabular}
\end{table*}

\subsection{Results of SSG-VQA-Net}
\label{sec:vqa_result}

\subsubsection{Results on SSG-VQA}

\textbf{Comparison to other works}. As shown in Table~\ref{table:overall}, SSG-VQA-Net outperforms baseline models like VisualBert ~\citep{visualbert} and VisualBert Resmlp ~\citep{seenivasan2022surgical} in metrics such as mAP, Recall, and Fscore. We also train an upper-bound model, called SSG-VQA-Net (oracle), that uses the scene embeddings from detection labels of the SSG-VQA dataset instead of using the trained YOLOv7 object detector. This model outperforms prior works significantly, emphasizing the importance of high-quality scene embedding inputs.

\textbf{Analysis by question type}. As shown in Table~\ref{table:ssgqa_breakdown}, SSG-VQA-Net can handle various question types. For ``counting'' questions, it outperforms VisualBert by $7.3$ points in mAP. For ``existence'' and ``query object'' types, the model again shows superior performance w.r.t to baseline models. 

\textbf{Analysis by complexity}. SSG-VQA dataset provides the diagnostic setup to pinpoint the weakness of the model. As shown in Table~\ref{table:ssgqa_breakdown}, we compute the performance of our models on questions that require different visual reasoning complexity, i.e., \textit{zero-hop} and \textit{one-hop}, and \textit{single-and}. Our model shows consistent gains in both simple and complex question queries. For \textit{one-hop} and \textit{single-and} type of questions, SSG-VQA-Net achieves a $4.1$ and $8.1$ point mAP increase over VisualBERT ResMLP, respectively. This indicates that the inclusion of scene context can aid in resolving complex queries.

\thispagestyle{default}
\subsubsection{Results on Cholec80-VQA}

\begin{table*}[t!]
\centering
\caption{Results on the Cholec80-VQA. SSG-VQA-Net achieves better results than the state-of-the-art models, even w.r.t SurgicalGPT, which contains a heavy sequence decoding module of GPT-2}\label{table:cholec80}
\begin{tabular}
{c@{\hspace{1cm}}c@{\hspace{1cm}}c@{\hspace{1cm}}c@{\hspace{1cm}}c@{\hspace{1cm}}c} 
\toprule
  & \#Parameter    & Accuracy & Recall & Fscore  \\
\midrule
VisualBert ~\citep{visualbert} & 209.8M & 89.7 & 62.9 & 63.3 \\ 
VisualBert Resmlp ~\citep{seenivasan2022surgical} & 184.7M & 89.8 & 62.7 & 63.4 \\
SurgicalGPT ~\citep{seenivasan2023surgicalgpt} & 234.5M & 87.5 & 57.5 & 57.9\\
SSG-VQA-Net & 145.3M & \textbf{90.6} & \textbf{64.4} & \textbf{63.7}\\ 
\bottomrule
\end{tabular}
\end{table*}

We also conduct the experiments on the other publicly available surgical VQA dataset Cholec80-VQA. As illustrated in Table~\ref{table:cholec80}, SSG-VQA-Net significantly outperforms the SurgicalGPT ~\citep{seenivasan2023surgicalgpt}, which requires heavy sequence decoding using GPT-2 architecture. This highlights that the bottleneck of the current surgical VQA problem lies in the visual scene understanding instead of text generation. Also, even using YOLOv7 for object detection, our model achieves higher performance with fewer parameters than prior works, verifying its efficiency while maintaining higher performance metrics.

\subsubsection{Ablation study}

\begin{table*}[t!]
\centering
\caption{Effect of different modules. RoIAlign pooling boosts results, and the Scene-embedded Interaction Module further enhances them. Both modules offer complementary benefits.}\label{table:ablation}
\begin{tabular}
{c@{\hspace{1.2cm}}c@{\hspace{1.2cm}}c@{\hspace{1.2cm}}c@{\hspace{1.2cm}}c@{\hspace{1.2cm}}c@{\hspace{1.2cm}}c} 
\toprule
SIM & ROI & Accuracy &mAP & Recall & Fscore\\
\midrule
$\times$ & $\times$ & 55.0 & 47.9 & 42.5 & 54.8 \\
$\checkmark$ & $\times$ & 57.1 & 51.0 & 47.2 & 56.1 \\
$\times$ & $\checkmark$ & 58.8 & 52.5 & 47.9 & 57.7 \\ 
$\checkmark$ & $\checkmark$ & \textbf{60.7} & \textbf{54.9} & \textbf{49.1} & \textbf{60.3} \\ 
\bottomrule
\end{tabular}
\end{table*}

Table~\ref{table:ablation} shows that combining both the Scene-embedded Interaction Module (SIM) and RoIAlign (ROI) pooling significantly boosts the model's performance. This suggests that these modules are not just individually beneficial but are actually complementary. Specifically, the model attains the highest mAP (54.9\%) when both components are added. Also, the improvement indicates that introducing scene knowledge representation learning is crucial for robust surgical visual question answering. 

\thispagestyle{default}
\section{Conclusion}

In this paper, we tackle the problem of visual question answering (VQA) in the context of fine-grained surgical scene understanding. First, we introduce a new dataset called SSG-VQA, which uses a surgical scene graph as an underlying representation and a question-answer generation engine to generate diverse, geometrically grounded, and surgical-action-oriented question-answer pairs. The question-answer pairs are also sampled to mitigate the question-condition bias that exists in the current surgical VQA datasets. We also propose a novel model called SSG-VQA-Net to explicitly incorporate scene knowledge and object-wise local features in the VQA model design to improve the reasoning ability on complex questions. The results show that SSG-VQA-Net outperforms existing baseline models by a large margin. 

\thispagestyle{default}

\section{Acknowledgments}
This work has received funding from the European Union (ERC, CompSURG, 101088553). Views and opinions expressed are however those of the authors only and do not necessarily reflect those of the European Union or the European Research Council. Neither the European Union nor the granting authority can be held responsible for them. This work was also partially supported by French state funds managed by the ANR under Grants ANR-20-CHIA-0029-01 and ANR-10-IAHU-02. This work was granted access to the HPC resources of IDRIS under the allocation AD011013704R1 and AD011011631R2 made by GENCI. This work was granted access to the HPC resources of IDRIS under the allocations AD011013704R1, AD011011631R2, and AD011011631R3 made by GENCI. The authors would like to acknowledge the High Performance Computing Center of the University of Strasbourg for supporting this work by providing scientific support and access to computing resources. Part of the computing resources were funded by the Equipex Equip@Meso project (Programme Investissements d'Avenir) and the CPER Alsacalcul/Big Data.

\bibliographystyle{model2-names.bst}
\bibliography{sn-bibliography}

\clearpage
\setcounter{page}{1}
\maketitlesupplementary
\thispagestyle{firstpagestyle}

\subsection{SSG Dataset}
\label{sec:full_training}

In this section, we first illustrate the statistics of the SSG-VQA dataset and show that it is more challenging and diverse than the prior works. Then, we show the process of assigning location attributes to the surgical objects. Finally, we visualize some samples of the scene graphs that we use to generate question-answer pairs.

\begin{table}[t!]
\centering
\caption{Dataset statistics comparison. SSG-VQA is more challenging and balanced as it includes more attributes and complexities in the questions.}
\label{table:dataset_stats}
\scalebox{0.80}{
    \begin{tabular}{cccc}
        \hline
        Dataset & EndoVis-18-VQA  & Cholec80-VQA & SSG-VQA  \\
        \hline
        Average Length & 5.8  & 2.0 & 12.8 \\
        Average $\#$Questions & 5.0 & 6.5 & 38.9\\
        \hline
        $\#$Color & 0 & 0 & 7\\
        $\#$Type & 0 & 0 & 2\\
        $\#$Location & 4 & 0 & 6\\
        $\#$Object & 9 & 7 & 13\\
        $\#$Relation & 13 & 0 & 16 \\
        \hline
    
    \end{tabular}
    }
\end{table}

\subsubsection{Dataset Statistics}
As shown in Tab.~\ref{table:dataset_stats}, SSG-VQA dataset contains the more challenging and diverse question-answer pairs. Specifically, the average length of question is $38.9$ words compared to $5$ words and $6.5$ words from EndoVis-18-VQA and Cholec80-VQA. Also, by using the question engine to automatically generate the question-answer pairs, SSG-VQA contains a more diverse and different pattern of question-answer pairs for each surgical scene.

\subsubsection{Location Attribute and Triplet Annotation}

\begin{figure}[htbp]
    \centering
    \includegraphics[width=1.0\linewidth]{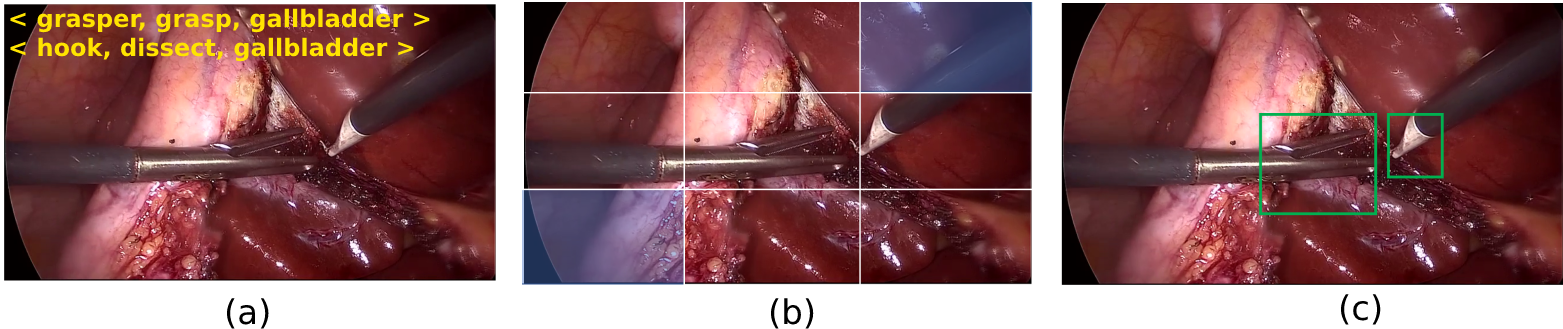}
    \caption{\textbf{SSG-VQA Dataset - location attribute:} (a) we use triplet annotations for action-related question-answer generation; (b) we include a spatial attribute for each object, e.g., top-right, bottom-left; (c) we use tool and anatomy bounding boxes to generate question-answer pairs.}
    \label{location_attribute}
\end{figure}

During the question-answer generation, we assign the location attributes to the detected surgical objects, including instruments and anatomies. As shown in Tab.~\ref{location_attribute}, we split the image into $9$ spatial blocks and assign the location attribute based on the locations of the surgical object in a given block. In addition to the spatial relations among surgical objects, we also use triplet annotation to connect the surgical objects with the actions, as shown in Fig. \ref{location_attribute} (a). 

\subsubsection{Scene Graphs}

\begin{figure}[t!]
    \centering
    \includegraphics[width=1.00\linewidth]{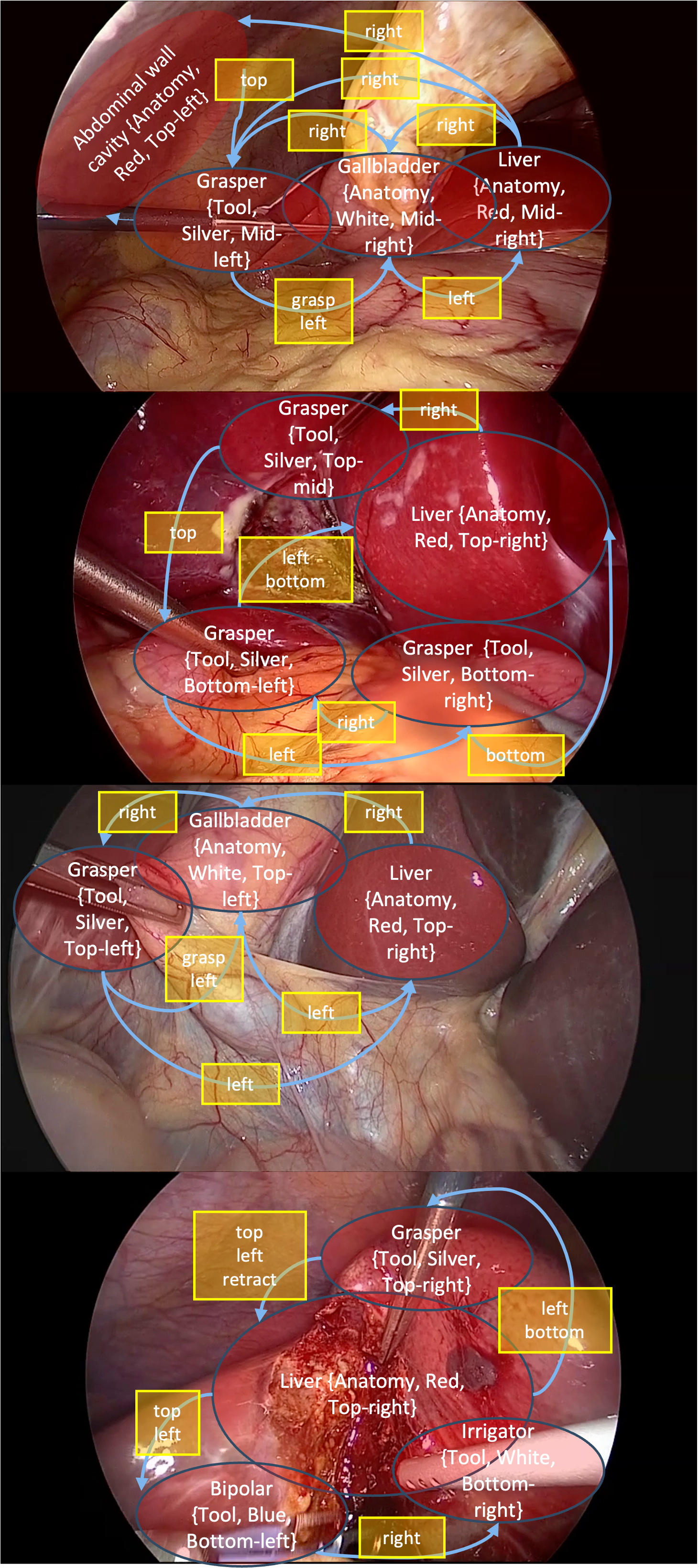}
    \caption{\textbf{SSG-VQA Dataset - scene graphs:} The questions are generated by exploiting the surgical scene graph. The surgical scene graph consists of spatial attributes and action relations.}
    \label{sample_scene}
\end{figure}

\begin{figure*}[t!]
    \centering
    \includegraphics[width=0.95\linewidth]{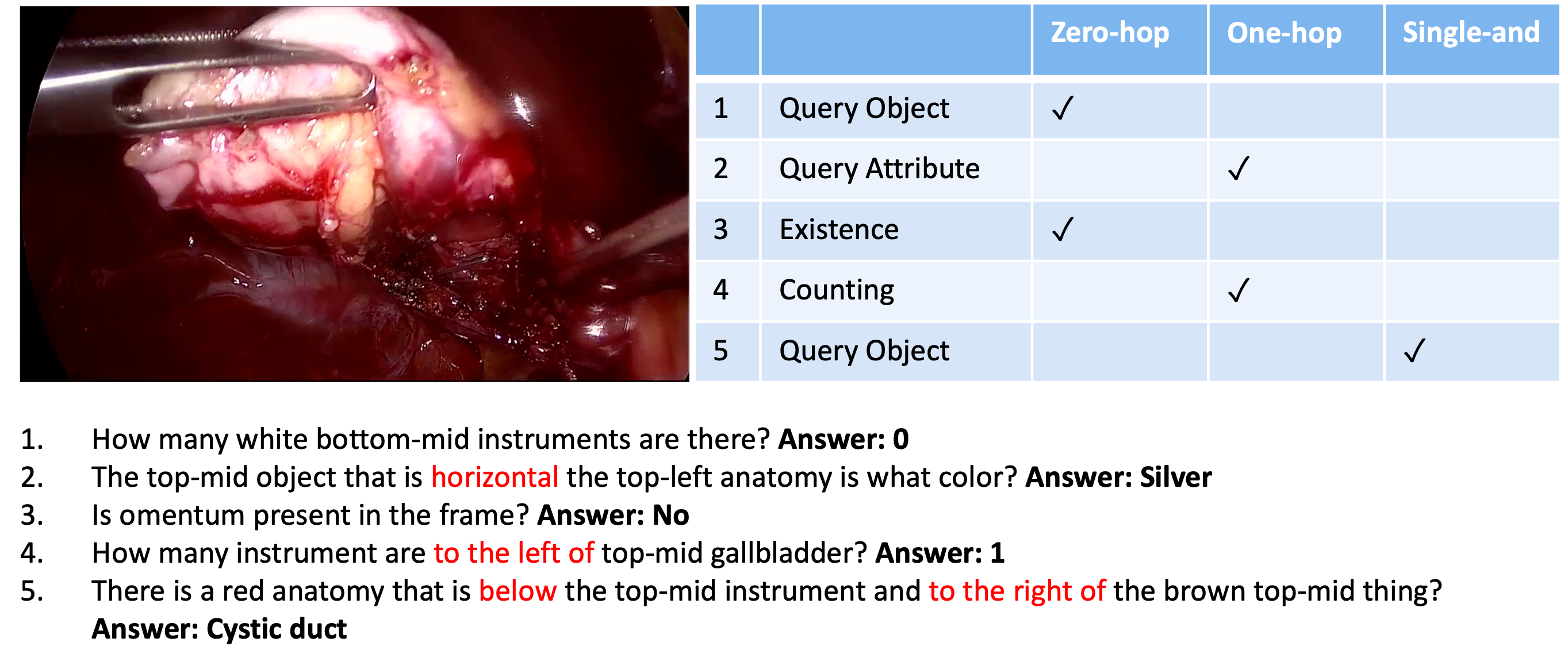}
    \caption{\textbf{SSG-VQA Dataset - different complexities:} We generate different type of question-answer pairs for each surgical scene image. The text in red indicates the reasoning step in the image, e.g., the fifth question requires two steps of reasoning to solve, therefore categorized into single-and type.}
    \label{complexity_qa}
\end{figure*}

\begin{figure*}[t!]
    \centering
    \includegraphics[width=0.95\linewidth]{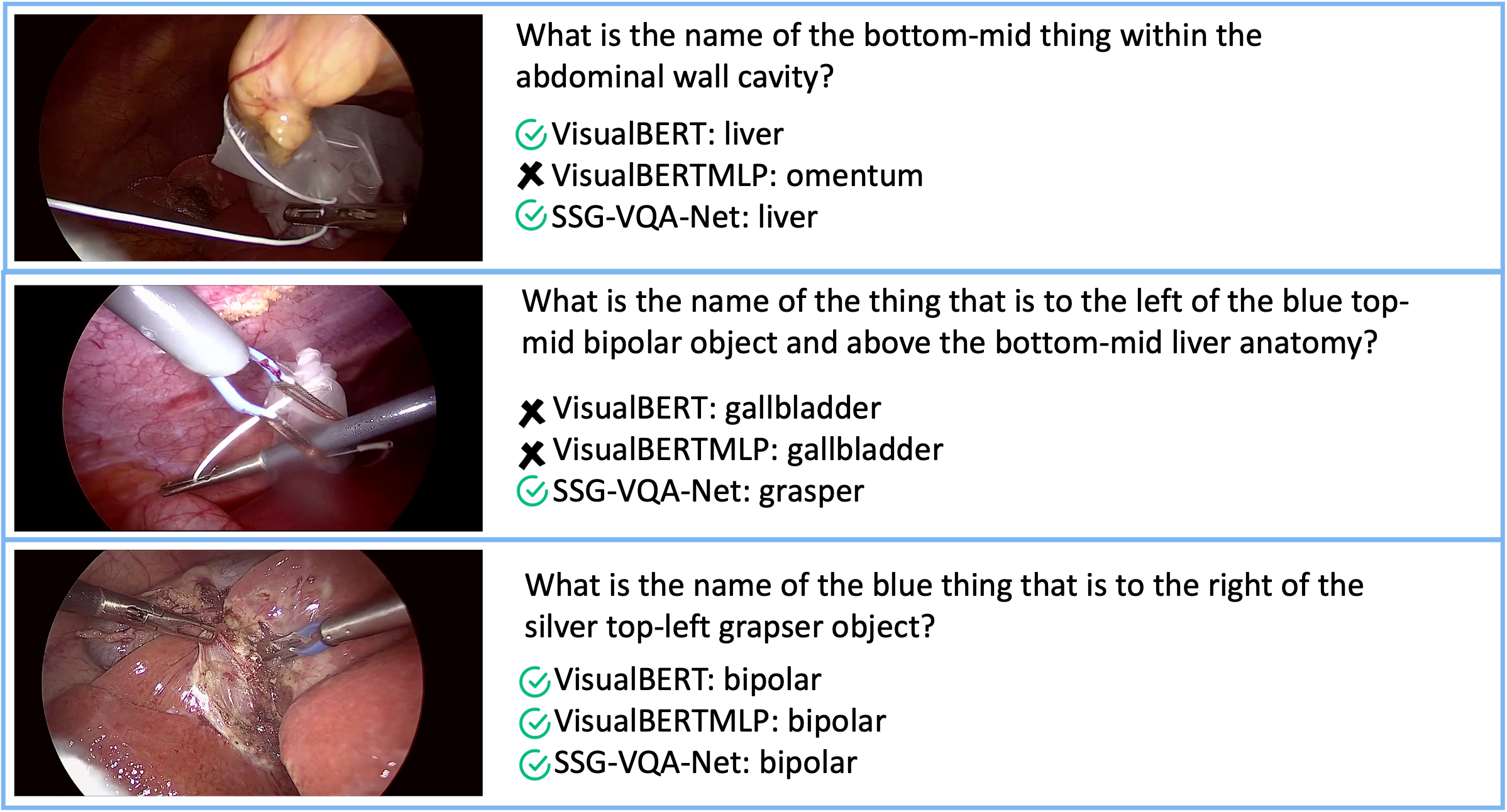} %
    \caption{\textbf{Qualitative Results:} Comparison of SSG-VQA-Net with baseline surgical VQA models on the question-answer pairs based on spatial, color, counting, anatomy, and tool presence attributes.}
    \label{qa:attributes}
\end{figure*}

In this section, we show some scene graph samples that we use to generate question-answer pairs. As shown in Fig. \ref{sample_scene}, the surgical semantic scene graphs are directed graphs where the nodes represent the surgical objects and the edges represent the spatial and action relation between nodes. Also, we assign attributes to each surgical object, e.g., type, color, and location of the objects. Note that the spatial relations are fully connected among the objects. We show the important relations for a simpler demonstration.

\subsubsection{Complexities of Question-answer Pairs}

As the SSG-VQA dataset is generated with a question engine and question templates automatically, we control the complexity of question-answer pairs by altering the question templates. The question-answer pairs with different complexities also offer the diagnostic ability to pinpoint the weakness of the VQA model. As shown in Fig. \ref{complexity_qa}, we have $5$ kinds of question-answer pairs with $3$ kinds of complexities. Here, the complexity is measured based on how many reasoning steps are required to answer the question. For example, the fifth question-answer in Fig. \ref{complexity_qa} requires two steps of reasoning to solve, highlighted in \textcolor{red}{red}. As a result, this question is more complex than the others.

\begin{figure*}[t!]
    \centering
    \includegraphics[width=0.95\linewidth]{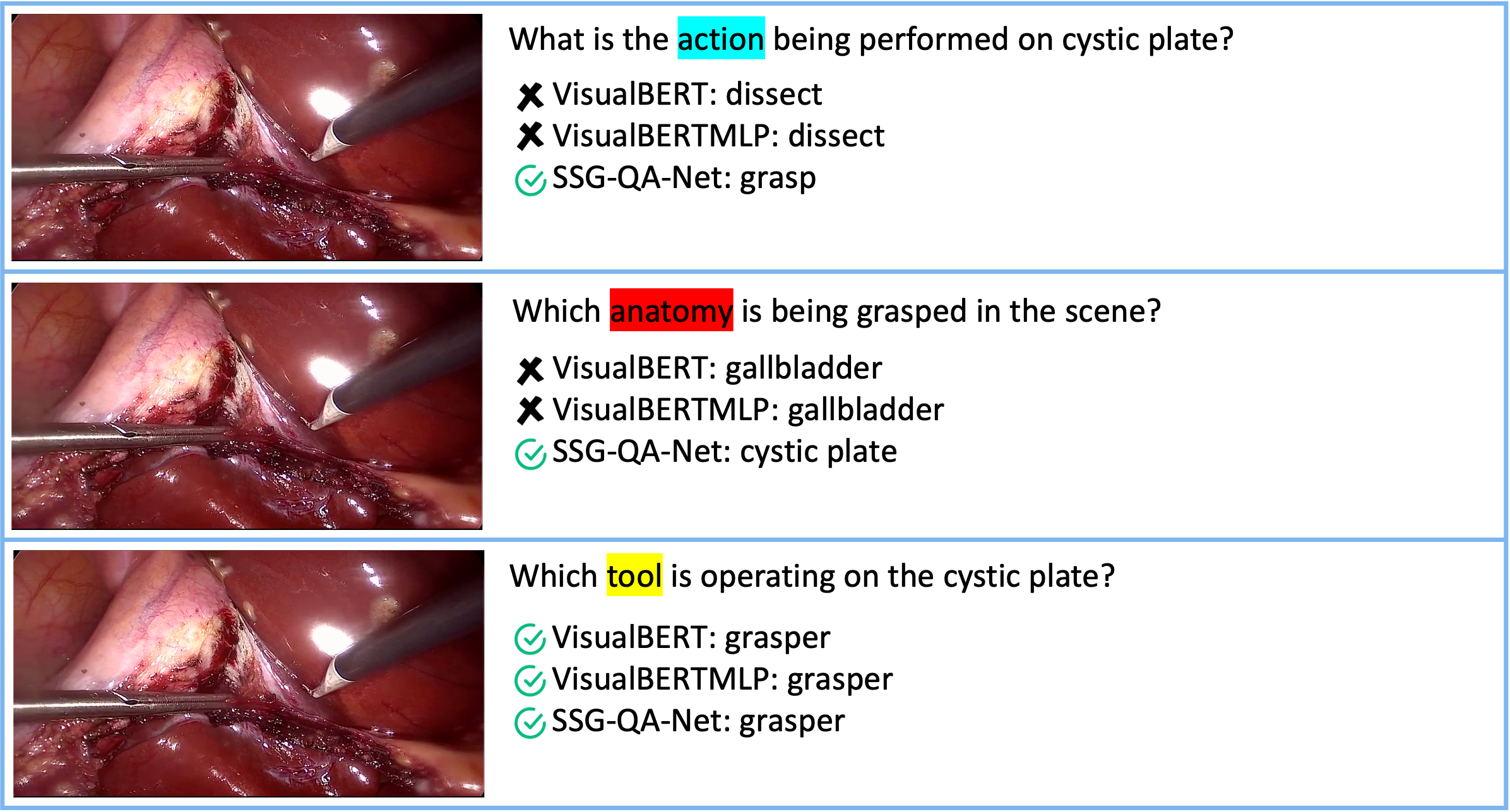}
    \caption{\textbf{Qualitative Results:} Comparison of SSG-VQA-Net with baseline surgical VQA models on the question-answer pairs based on surgical action triplets.}
    \label{qa:triplet}
\end{figure*}

\begin{figure*}[t!]
    \centering
    \includegraphics[width=0.95\textwidth]{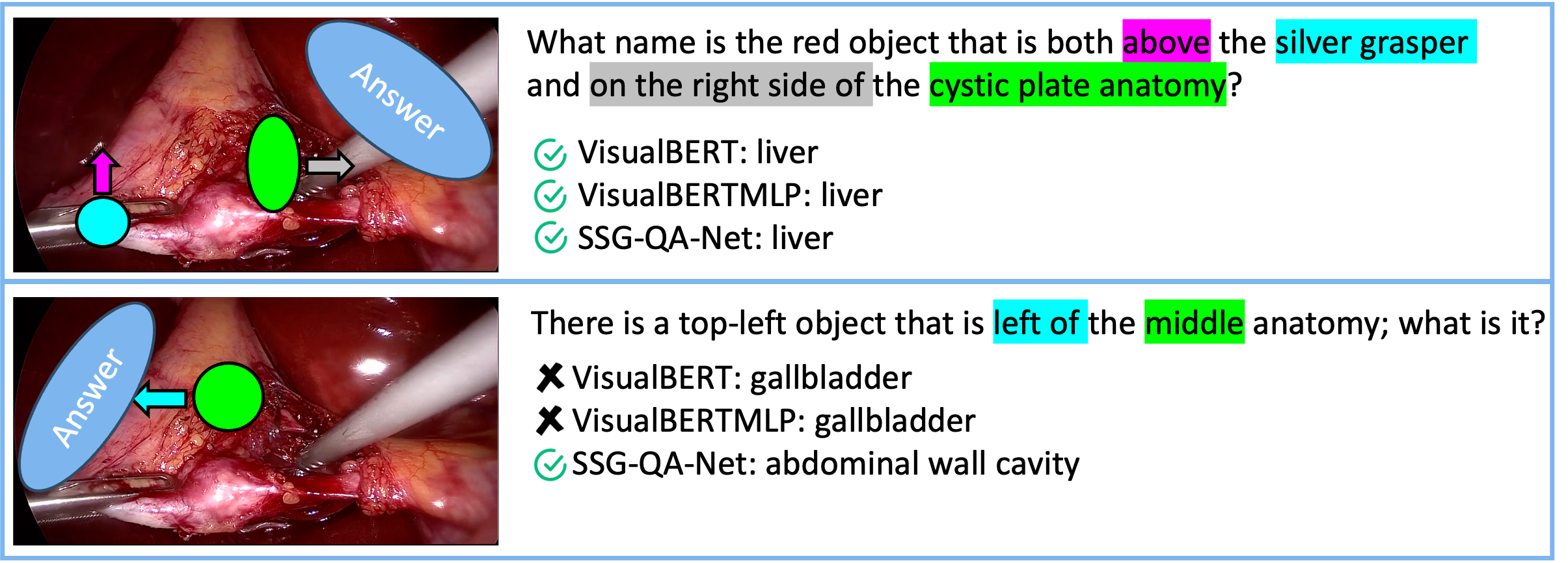} %
    \caption{\textbf{Qualitative Results:} It shows that the questions from our SSG-VQA dataset require visual reasoning to solve. Also, our SSG-VQA-Net with scene knowledge outperforms the prior baselines in terms of complex questions. }
    \label{qa:reasoning steps}
\end{figure*}

\thispagestyle{default}
\subsection{Qualitative Results}

In this section, we compare our SSG-VQA-Net to the state-of-the-art surgical VQA models in terms of the question that require the understanding of attributes (Sec. \ref{supp_sec:attribute}), action triplet occurrence (Sec. \ref{supp_sec:triplet}) and reasoning steps (Sec. \ref{supp_sec:reasoning}) to solve. 

\subsubsection{Attributes}
\label{supp_sec:attribute}

As shown in Fig.~\ref{qa:attributes}, the SSG-VQA-Net model outperforms the prior works when the questions contain attributes, such as location and color. For example, the second subfigure in Fig.~\ref{qa:attributes} shows that the SSG-VQA-Net model understands the location attributes such as ``top-mid'' and ``bottom-mid'' to resolve the question. Compared to the other methods, it contains more detailed knowledge about the surgical scene.

\subsubsection{Action Triplets}
\label{supp_sec:triplet}

We also compare the SSG-VQA-Net model to the prior works on the questions that require an understanding of surgical action triplet. Fig.~\ref{qa:triplet} shows that the SSG-VQA-Net model can answer the questions that are generated based on the triplet annotations. 

\thispagestyle{default}
\subsubsection{Reasoning Steps}
\label{supp_sec:reasoning}

Fig.~\ref{qa:reasoning steps} shows that SSG-VQA-Net can perform multiple steps of reasoning to generate the correct answer to complex questions. Compared to the prior works, it achieves better results when the questions are more complex.

\end{document}